\documentclass[11pt]{article}

\usepackage[utf8]{inputenc}
\usepackage[T1]{fontenc}
\usepackage[margin=1in]{geometry}
\usepackage{times}
\usepackage{microtype}
\usepackage{graphicx}
\usepackage{booktabs}
\usepackage{amsmath, amssymb}
\usepackage{hyperref}
\usepackage{enumitem}
\usepackage{xcolor}
\usepackage{caption}
\usepackage{array}
\usepackage{listings}
\usepackage{fancyvrb}

\hypersetup{
    colorlinks=true,
    linkcolor=black,
    citecolor=black,
    urlcolor=blue
}

\title{\textbf{Operand Quant: A Single-Agent Architecture for Autonomous Machine Learning Engineering}}
\author{
\textbf{Operand Research} \\
Arjun Sahney, Ram Gorthi, Cezary Łastowski, Javier Vega
}
\date{October 2025}

\begin{document}
\maketitle

\begin{abstract}
We present \textbf{Operand Quant}, a single-agent, IDE-based architecture for autonomous machine learning engineering (MLE). Operand Quant departs from conventional multi-agent orchestration frameworks by consolidating all MLE lifecycle stages---exploration, modeling, experimentation, and deployment---within a single, context-aware agent. On the MLE-Benchmark (2025), Operand Quant achieved a new state-of-the-art (SOTA) result, with an overall medal rate of \textbf{0.3956 $\pm$ 0.0565 across 75 problems}---\textbf{the highest recorded performance among all evaluated systems to date}. The architecture demonstrates that a linear, non-blocking agent, operating autonomously within a controlled IDE environment, can outperform multi-agent and orchestrated systems under identical constraints.
\end{abstract}

\section{Introduction}
The automation of the machine learning engineering (MLE) pipeline has become a central objective in agentic AI research. Many recent systems rely on multi-agent orchestration, in which specialized agents independently handle data analysis, modeling, evaluation, and deployment. While such frameworks can parallelize work, they often incur coordination costs, context fragmentation, and synchronization errors.

\textbf{Operand Quant} explores an alternative paradigm: a single autonomous agent that continuously observes, plans, edits, executes, and evaluates within its own integrated development environment (IDE). This design hypothesis asserts that end-to-end contextual continuity can yield reliable and efficient performance without distributed orchestration.

The agent was evaluated on the \textbf{MLE-Benchmark} \cite{mlebench}, which tests autonomous ML capabilities under isolation (no internet, fixed runtime, standardized hardware). Operand Quant's results establish that a single-agent architecture can achieve top-tier performance on complex machine learning engineering tasks without relying on external retrieval or multi-agent decomposition.

\section{Related Work}

\subsection{Multi-Agent Systems for Machine Learning Experimentation}
A growing line of work investigates \emph{multi-agent} pipelines for end-to-end machine learning experimentation (MLE). AutoML-GPT-style systems couple an LLM planner with tool-augmented executors to select models, features, and hyperparameters automatically \cite{automlgpt-zhang-2023,automlgpt-tsai-2023}. More recently, \emph{AutoML-Agent} proposes a specialized ensemble of agents spanning data acquisition to deployment, with retrieval-augmented planning and explicit role decomposition; it reports competitive end-to-end results on public tasks and featured at ICML 2025 \cite{automlagent}. Complementary to systems papers, \textbf{MLAgentBench} formalizes tasks where agents must run actual ML experiments (edit code, execute training, iterate), enabling controlled comparisons across agent architectures \cite{mlagentbench}.

Relative to these works, \textbf{Operand Quant} takes the opposite tack: it eschews role-based decomposition and instead maintains a \emph{single} persistent IDE-resident agent with unified state. This removes cross-agent handoffs and synchronization policies that can fragment context or incur orchestration overhead. Our evaluation on MLE-Benchmark (governed, offline) complements prior open-world and online-tool settings emphasized by many multi-agent systems.

\subsection{Single-Agent Coding Systems}
Single-agent coding agents show that an LLM equipped with a rich "computer interface" can solve complex software tasks. \textbf{SWE-agent} introduces an Agent–Computer Interface (ACI) that enables repository-scale navigation, editing, and execution; it achieves SOTA on SWE-bench and HumanEvalFix under interactive settings \cite{sweagent}. In parallel, code-specialized foundation models (e.g., \textbf{CodeT5}, \textbf{CodeT5+}) improve editing/generation quality via identifier-aware pretraining and unified encoder–decoder objectives \cite{codet5,codet5plus}.

Operand Quant aligns more with the single-agent ACI paradigm than with multi-agent orchestration: the agent continuously observes an IDE-like workspace, plans actions, edits/runs code, and evaluates outcomes. However, unlike many coding agents that optimize for unit-test pass rates or repository repairs, Operand Quant targets the \emph{MLE lifecycle} (EDA, training, evaluation, iteration, and submission) under benchmark governance.

\subsection{Traditional AutoML Approaches}
Classic AutoML frameworks such as \textbf{AutoGluon} and \textbf{H2O AutoML} automate model selection and hyperparameter tuning primarily through ensembling, stacking, and search strategies; they require minimal code but do not reason about open-ended editing or end-to-end project structure \cite{autogluon,h2oautoml}. AutoGluon's multi-layer stack ensembling and H2O's fast random search with stacked ensembles establish strong tabular baselines that remain competitive in constrained settings.

Operand Quant differs conceptually: rather than exposing a fixed "fit-and-tune" API, it operates as an \emph{autonomous engineer} that writes code, configures experiments, interprets logs, and adapts pipelines. Traditional AutoML can serve as a tool within such agents, but it does not substitute for workspace-level planning and iterative development.

\subsection{Agentic AI Frameworks}
General-purpose agent frameworks provide the building blocks for orchestration, memory, tools, and long-running control. \textbf{LangGraph} emphasizes stateful, long-lived agents and graph-structured control flow \cite{langgraph}; \textbf{AutoGen}/AG2 offers multi-agent conversation patterns and event-driven workflows \cite{autogen,autogen-docs}; \textbf{CrewAI} focuses on role-based multi-agent "crews" with guardrails and observability \cite{crewai,crewai-docs}; \textbf{OpenAI Swarm} explores lightweight, educational handoffs among agents \cite{swarm}. \textbf{LlamaIndex} supplies agent abstractions and workflows tightly integrated with retrieval and tool use \cite{llamaindex-agents,llamaindex-overview}.

Operand Quant is orthogonal to these frameworks: it is an \emph{architecture and evaluation} of a single-agent IDE-centric system under offline benchmark constraints, rather than a general orchestration library. Our results suggest that when the task distribution matches MLE-Benchmark's governed setup, preserving a unified, non-blocking reasoning state can out-perform role-decomposed multi-agent stacks.

\paragraph{Summary.}
Multi-agent MLE systems highlight breadth via specialization and retrieval; single-agent coding systems show depth via tight computer interfaces; traditional AutoML optimizes fixed pipelines; and agentic frameworks provide orchestration primitives. Operand Quant demonstrates that a single, context-continuous agent operating inside an IDE can deliver SOTA on governed MLE tasks without multi-agent coordination.

\section{System Overview}

\subsection{Design Principles}
Operand Quant is built as a single-agent system embedded within a simulated IDE. The IDE emulates the tools of a human ML engineer---file explorer, notebooks, scripts, execution kernels, and logs---allowing the agent to autonomously perform all phases of the MLE lifecycle: exploratory data analysis, feature engineering, modeling, evaluation, iteration, and productionization.

In contrast to orchestrated multi-agent setups, Operand Quant maintains a \emph{unified reasoning state} throughout execution. The same LLM instance plans, codes, executes, and evaluates, eliminating context handoffs and preserving a consistent view of the workspace.

\subsection{Non-Blocking Turn-Based Operation}
The agent operates in \emph{turns}, each representing one reasoning--execution cycle. During a turn: (i) the agent observes current IDE state (open files, kernel status, active processes, and outputs); (ii) decides an action and emits a single structured JSON command; (iii) the action is validated and executed asynchronously; and (iv) the result is persisted and integrated into history. If an execution process remains active, the agent continues working on other tasks while monitoring intermediate outputs. This non-blocking loop enables parallel reasoning and continuous iteration.

\section{Deep-Thinking Ensemble}

\subsection{Motivation}
Large language models (LLMs) exhibit context bias, a degradation of reasoning flexibility as prompt length increases. In long reasoning sessions, models can develop tunnel vision, reducing their ability to debug or reassess prior assumptions. Operand Quant mitigates this limitation via a mechanism called \emph{deep-thinking}.

\subsection{Ensemble Reasoning}
When the agent encounters a reasoning bottleneck, it delegates the problem to an ensemble of high-capacity models---\textbf{GPT-5}, \textbf{Claude-4.1 Opus}, \textbf{Grok-4}, and \textbf{Gemini 2.5 Pro}---that independently generate analyses or hypotheses. Their outputs are synthesized into a consolidated ``expert review,'' reintroduced into the agent's reasoning context as advisory input. From the agent's perspective, this appears as a consultation with domain experts; in practice, it constitutes a structured ensemble reasoning step performed under strict isolation.

\subsection{Compliance}
All ensemble models operated without internet access, satisfying MLE-Benchmark requirements. Ensemble interactions were limited to local inference; no external retrieval or tool augmentation was used.

\section{Architecture Specification}

\subsection{Core Agent Loop}
Each reasoning cycle proceeds as follows: (1) observe IDE and process state; (2) generate a structured JSON action conforming to a validated schema; (3) execute the specified operation; (4) persist results to disk; and (5) trigger compaction if nearing context length limits. A strict \emph{single-tool-per-turn} rule enforces interpretability and deterministic replay.

\subsection{Concurrent Execution}
Operand Quant supports asynchronous notebook and script execution. When a process is running, it remains under continuous monitoring for status, output, and resource utilization. Example monitoring logic:

\begin{lstlisting}[language=Python, caption={Asynchronous execution monitoring logic}]
if primary_notebook and primary_notebook.is_cell_executing():
    continue_result = primary_notebook.continue_execution_if_running()
    if continue_result["status"] == "completed":
        final_output = continue_result.get("output", "[No Output]")
    elif continue_result["status"] == "still_executing":
        current_output = continue_result["current_output"]
        duration = continue_result["execution_duration_seconds"]
\end{lstlisting}

This enables non-blocking concurrent processing: while training runs execute, the agent continues editing, planning, or analyzing outputs.

\subsection{Interruption Logic}
Execution processes are interrupted when: (i) convergence is detected from loss or validation metrics; (ii) memory or runtime thresholds are exceeded; or (iii) non-convergent patterns appear in logs or errors. This dynamic interruption mechanism allows efficient allocation of the fixed runtime budget.

\subsection{State Persistence and Compaction}
All process metadata, outputs, and dependencies are stored persistently. When cumulative context approaches model limits, Operand Quant performs hierarchical memory compaction: (1) exclude verbose notebook content; (2) summarize older turns using a dedicated summarization utility; (3) validate the summary; and (4) replace the original history upon successful validation. Each compaction prompt and output is saved in \texttt{agent\_metadata/} for auditability.

\subsection{IDE Context Awareness}
The agent receives a real-time textual summary of its environment, e.g.:
\begin{lstlisting}[caption={Real-time IDE environment status}]
Execution Status:
Cell 3 of model_training.ipynb executing (127 s)
validation_script.py running (45 s)
hyperparameter_search.ipynb idle
\end{lstlisting}
This enables reasoning about dependencies, parallelism, and resources with fine granularity.

\section{Benchmark Governance and Evaluation Protocol}

\subsection{Compliance}
Operand Quant fully complied with MLE-Benchmark 2025 governance: (i) no internet or API access; (ii) tools confined to local environment; and (iii) standardized submission via the \texttt{submit\_final\_answer} endpoint. An earlier alias, \texttt{submit\_for\_scoring}, was present in logs but non-functional; no run received live score feedback. All runs were verified through manual log inspection.

\subsection{Infrastructure}
Each run was limited to a 24-hour execution window. The \emph{Lite} subset used a GCP VM (234 GB RAM, 36 vCPUs, Tesla T4). The \emph{Medium/Hard} subsets used Azure NV36AdsA10v5 (official MLE hardware). Hardware transition between providers accounts for timing differences across runs.

\subsection{Submission Lifecycle}
When the agent achieved a medal result, it called \texttt{submit\_final\_answer}, synchronized code and metadata to the host, and terminated the container. If submission validation failed or no result was produced, the run was labeled ``no medal.'' Logs were preserved for reproducibility.

\section{Experimental Results}

Operand Quant followed the exact canonical MLE-Benchmark setup in terms of runtime and machine specifications, using the standard 24-hour runtime limit and NVAds36v4A10 machine build to ensure fair comparison with other evaluated systems. Operand Quant's performance was independently verified by the OpenAI Benchmark team.

\subsection{Performance Summary}
Operand Quant achieved the following medal rates on MLE-Benchmark 2025:
\begin{table}[h]
\centering
\caption{MLE-Benchmark Medal Rates}
\begin{tabular}{lcc}
\toprule
\textbf{Subset} & \textbf{Medal Rate (mean $\pm$ std)} & \textbf{Problems (n)} \\
\midrule
Overall & \textbf{0.3956 $\pm$ 0.0565} & 75 \\
Lite & \textbf{0.6364 $\pm$ 0.1050} & 22 \\
Medium & \textbf{0.3333 $\pm$ 0.0765} & 38 \\
Hard & \textbf{0.2000 $\pm$ 0.1069} & 15 \\
\bottomrule
\end{tabular}
\end{table}

\subsection{Leaderboard Comparison}
\begin{table}[h]
\centering
\caption{Leaderboard (2025-09)}
\small
\begin{tabular}{p{3.5cm}cccccc}
\toprule
\textbf{Agent} & \textbf{Lite} & \textbf{Med.} & \textbf{Hard} & \textbf{All} & \textbf{Hrs} & \textbf{Date} \\
\midrule
\textbf{Operand Quant} & \textbf{63.64} & \textbf{33.33} & \textbf{20.00} & \textbf{39.56} & \textbf{24} & \textbf{09-28} \\
InternAgent (DeepSeek-R1) & $62.12$ & $26.32$ & $24.44$ & $36.44$ & 12 & 09-12 \\
R\&D-Agent (GPT-5) & $68.18$ & $21.05$ & $22.22$ & $35.11$ & 12 & 09-26 \\
Neo Multi-Agent & $48.48$ & $29.82$ & $24.44$ & $34.22$ & 36 & 07-28 \\
R\&D-Agent (o3 + GPT-4.1) & $51.52$ & $19.30$ & $26.67$ & $30.22$ & 24 & 08-15 \\
\bottomrule
\end{tabular}
\end{table}

\noindent Operand Quant's 24-hour runs yield a higher overall score than all other published agents, including multi-agent architectures with shorter runtimes.

\subsection{Incomplete or Failed Problems}
The following tasks failed due to data or environment issues and are reported as ``no medal'' across seeds: 3D Object Detection for Autonomous Vehicles; AI4Code; Billion Word Imputation; BMS Molecular Translation; Google Research Identify Contrails; HMS Harmful Brain Activity Classification; HuBMAP Kidney Segmentation; Jigsaw Unintended Bias Classification; RSNA-MICCAI Brain Tumor Radiogenomic Classification; Statoil Iceberg Classifier; TensorFlow2 Question Answering. One outlier---Multi-Modal Gesture Recognition---was excluded after identifying a dataset leakage bug that led to an invalid perfect score.

\section{Discussion and Limitations}
Operand Quant's results demonstrate that a single-agent, turn-based architecture can outperform orchestrated systems on the MLE-Benchmark. Unified contextual reasoning and deterministic state persistence appear sufficient for competitive performance without distributed coordination. Limitations include: (i) context degradation despite compaction; (ii) one-tool-per-turn expressiveness limits; (iii) high compute cost from 24-hour runs; and (iv) incomplete fault tolerance for environment or kernel errors.

\section{Reproducibility and Logging}
Deterministic replay is enabled via comprehensive logging: reasoning turns and outcomes recorded in \texttt{full\_history.json}; IDE snapshots captured per turn in \texttt{IDE\_state.txt}; compaction prompts and summaries archived in \texttt{agent\_metadata/}; and figures generated by \texttt{scripts/generate\_arch\_\allowbreak diagrams.py}.

\section{Code and Data Availability}
All experimental artifacts, logs, and submission materials from the MLE-Benchmark evaluation are publicly available in the OperandLinear-MLE-Bench repository \cite{operandrepo}. The repository contains complete turn-by-turn execution logs (\texttt{full\_history.json}), all generated notebooks and scripts, verification scripts for medal rate calculations, environment setup configurations, and architecture specifications. This comprehensive logging enables full reproducibility of the reported results.

\section{Conclusion}
Operand Quant establishes a new state-of-the-art in autonomous machine learning engineering. With an overall score of \textbf{0.3956 $\pm$ 0.0565}, it currently ranks first on the MLE-Benchmark 2025 leaderboard, surpassing both single- and multi-agent baselines under identical governance conditions. Its success demonstrates that autonomous MLE systems can achieve leading performance using a unified, single-agent architecture grounded in continuous reasoning, concurrent execution, and structured context management. Future work will extend Operand Quant with adaptive ensemble reasoning, dynamic compaction, and fault-tolerant execution.

\end{document}